\documentclass[sigconf]{acmart}

\usepackage{balance}

\usepackage{hyperref}
\usepackage{cleveref}
\usepackage{graphicx}
\usepackage{enumitem}
\usepackage{multirow}
\usepackage{array}
\usepackage{subcaption}
\usepackage{microtype}
\usepackage{pdfpages}
\usepackage[symbol]{footmisc}

\def\BibTeX{{\rm B\kern-.05em{\sc i\kern-.025em b}\kern-.08emT\kern-.1667em\lower.7ex\hbox{E}\kern-.125emX}}
    
\copyrightyear{2019} 
\acmYear{2019} 
\setcopyright{rightsretained} 
\acmConference[KDD '19]{The 25th ACM SIGKDD Conference on Knowledge Discovery and Data Mining}{August 4--8, 2019}{Anchorage, AK, USA}
\acmBooktitle{The 25th ACM SIGKDD Conference on Knowledge Discovery and Data Mining (KDD '19), August 4--8, 2019, Anchorage, AK, USA}
\acmDOI{10.1145/3292500.3330723}
\acmISBN{978-1-4503-6201-6/19/08}

\begin{document}

%
\title{Gmail Smart Compose: Real-Time Assisted Writing}

%


\author{Mia Xu Chen}
\authornote{Equal contribution.}
\email{miachen@google.com}
\affiliation{\institution{Google}}

\author{Benjamin N Lee}
\authornotemark[1]
\email{blee@google.com}
\affiliation{\institution{Google}}

\author{Gagan Bansal}
\authornotemark[1]
\email{gaganbansal@google.com}
\affiliation{\institution{Google}}

\author{Yuan Cao}
\email{yuancao@google.com}
\affiliation{\institution{Google}}

\author{Shuyuan Zhang}
\email{syzhang@google.com}
\affiliation{\institution{Google}}

\author{Justin Lu}
\email{jtlu@google.com}
\affiliation{\institution{Google}}

\author{Jackie Tsay}
\email{jackietsay@google.com}
\affiliation{\institution{Google}}

\author{Yinan Wang}
\email{ynwang@google.com}
\affiliation{\institution{Google}}

\author{Andrew M. Dai}
\email{adai@google.com}
\affiliation{\institution{Google}}

\author{Zhifeng Chen}
\email{zhifengc@google.com}
\affiliation{\institution{Google}}

\author{Timothy Sohn}
\email{tsohn@google.com}
\affiliation{\institution{Google}}

\author{Yonghui Wu}
\email{yonghui@google.com}
\affiliation{\institution{Google}}

%
\renewcommand{\shortauthors}{Chen, Lee, Bansal, et al.}

%
\begin{abstract}


In this paper, we present Smart Compose, a novel system for generating interactive, real-time suggestions in Gmail that assists users in writing mails by reducing repetitive typing. In the design and deployment of such a large-scale and complicated system, we faced several challenges including model selection, performance evaluation, serving and other practical issues. At the core of Smart Compose is a large-scale neural language model. We leveraged state-of-the-art machine learning techniques for language model training which enabled high-quality suggestion prediction, and constructed novel serving infrastructure for high-throughput and real-time inference. Experimental results show the effectiveness of our proposed system design and deployment approach. This system is currently being served in Gmail.

\end{abstract}

%

%
\begin{teaserfigure}
\centering
\includegraphics[width=0.8\textwidth]{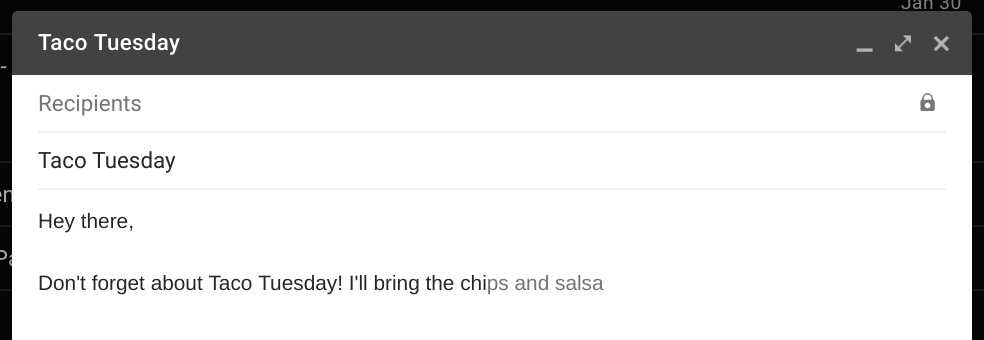}
\caption{Smart Compose Screenshot.}
\label{fig:teaser}
\end{teaserfigure}

%
\maketitle

\section{Introduction}

E-mail continues to be a ubiquitous and growing form of communication all over the world, with an estimated 3.8 billion users sending 281 billion e-mails daily \cite{radicati}. Improving the user experience by simplifying the writing process is a top priority of major e-mail service providers like Gmail. To fulfill this goal, previously Gmail introduced Smart Reply \cite{Kannan16}, a system for automatically generating short replies in response to incoming e-mail messages. While such a feature significantly reduces user response time, the suggestions are constrained to short phrases and appear only once in the composition process. Even with an initial suggestion, drafting longer messages can still be a time-consuming process, and is arguably one in which the user most needs accurate and frequent suggestions.

In this paper, we introduce Smart Compose, a system for providing real-time, interactive suggestions to help users compose messages quickly and with confidence in Gmail. Smart Compose helps by cutting back on repetitive idiomatic writing via providing immediate context-dependent suggestions. Unlike Smart Reply, Smart Compose assists with composing new messages from scratch and provides much richer and more diverse suggestions along the way, making e-mail writing a more delightful experience.
Each week, the system saves users over one billion characters of typing.

At the core of Smart Compose is a powerful neural language model trained on a large amount of e-mail data. The system makes instant predictions as users type. To provide high quality suggestions and a smooth user experience, we need to properly handle a variety of issues including model evaluation and large-scale inference, which we detail below.


\subsection{Challenges}
In creating Smart Compose, we faced several challenges not considered by previous work.
\begin{itemize}
\item \emph{Latency}. A key difference between Smart Compose and the previously described Smart Reply system is its real-time nature. Since suggestions appear as the user is typing, minimizing end-to-end latency is critical. The system requires the 90th percentile latency to be under 60ms.\footnote{According to~\cite{Nielsen93}, 0.1 second is about the limit for having the user feel that the system is reacting instantaneously.} Model inference needs to be performed on almost every keystroke, which can be particularly challenging for neural language models that are computationally expensive and not easily parallelized.
\item \emph{Scale}. Gmail is used by more than 1.5 billion diverse users. In order to produce suggestions that will be useful to most users, the model needs to have enough  capacity so that it is able to make tailored, high-quality suggestions in subtly different contexts.
\item \emph{Personalization}. Users often have their own unique e-mail writing styles. In order to make Smart Compose suggestions more similar to what the users would normally write, the new system needs to capture the uniqueness of their personal style.
\item \emph{Fairness and Privacy}.
In developing Smart Compose, we need to address sources of potential bias in the training process, and have to adhere to the same rigorous user privacy standards as Smart Reply, making sure that our models never expose user's private information. Furthermore, we had to develop this system without anyone on the project being able to look at the underlying data.

\item \emph{Metrics Design}. It is expensive and inefficient to directly evaluate candidate models on real user groups in online settings, we therefore need to devise proper efficient offline evaluation strategy and metrics which correlate well with online metrics that are critical to the product.
\end{itemize}

We give a detailed account for each of the challenges listed above in the remaining of the paper. Section \ref{related_work} introduces related work, especially the application of language models in different tasks; Section \ref{find_model} describes data preparation procedures, model architecture and evaluation criteria; Section \ref{production} gives an account of the production system and how we handle practical issues for high-quality serving; Section \ref{personalization} and Section \ref{multilingual} describe two extensions to the main Smart Compose model: personalization and internationalization; Section \ref{fairness} addresses privacy and fairness concerns.

\section{Related Work} \label{related_work}
Smart Compose is an example of a large-scale language model application. Language modelling is a fundamental and indispensable component of many natural language processing (NLP) and automatic speech recognition (ASR) systems~\cite{Jelinek97,Manning99,Jurafsky00,Koehn11}. Traditionally language models were dominated by n-gram models and their variants. In recent years with the rapid development in deep learning techniques, a multitude of neural network-based language models have been proposed, significantly improving state-of-the-art performance~\cite{Bengio03,Mikolov10,Jozefowicz16,Melis17,Yang18,Zolna18,Peters18}. In practice, language models are often integrated with upstream models to improve fluency of generated candidate word sequences. Language models alone can also be used in cases like input method to provide suggestions for subsequent words and to accelerate user typing (see for example~\cite{Goodman02,Gboard17}).

Despite the superior quality realized by neural language models, serving these models in large-scale brings a lot of challenges. The traditional n-gram language model permits easy factorization in probability computation and compact representation forms like finite state automaton~\cite{Mohri97}, which makes it amenable to large-scale distributed serving~\cite{Brants07,Emami07}. Neural language models in contrast, are stateful, heavy-weight and computationally expensive. By leveraging the latest deep learning model inference technologies~\cite{TPUv1, TPUv2}, our Smart Compose system gives one example of how neural language models can be served at industry-scale for billions of users.

Perhaps most related to the application of Smart Compose is Google's Smart Reply~\cite{Kannan16}. Smart Reply generates automated response suggestions for e-mails. It relies on a sequence-to-sequence model~\cite{Sutskever14} which first reads a message with an encoder, then outputs a distribution over a set of possible replies. Smart Compose differs from Smart Reply in several ways: First of all, Smart Reply only needs to generate responses once for each mail, whereas Smart Compose attempts to suggest possible extensions for every sentence prefix a user has typed in real time. Secondly, the possible responses given by Smart Reply is constrained to a relatively small human-curated whitelist with limited context-dependency, in contrast Smart Compose suggestions are more open-ended, flexible and sensitive to context. These differences are expected to improve the user mail-writing experience, but also bring about many challenges in model training, evaluation and serving.

The Smart Compose model is extended for personalization, as detailed in Section \ref{personalization}. As an important feature, personalization has been considered as a specific language model adaptation techniques in various applications, see for example~\cite{Hsu07,Tseng15,Lee2016,Yoon2017,Aaron2018}. In terms of the personalized model architecture, perhaps most close to our approach is~\citet{Chen15}, which is also a linear interpolation between an n-gram and recurrent neural network language model. Another extension of Smart Compose is the multilingual model as described in Section \ref{multilingual}. Multilingual models
are critical for internationalization, and are an active research topic in NLP, especially for tasks like machine translation, see for example~\cite{Orhan16,Melvin17,Lample19}. Our approach is much inspired by~\citet{Melvin17} in that we also consider using wordpieces and shared models for all languages.

\section{Finding the Right Model} \label{find_model}

The fundamental task in Smart Compose is to predict a sequence of tokens of variable length, conditioned on the prefix token sequence typed by a user and additional contextual information.
Similar to most sequence prediction tasks, during training, the objective is to maximize the log probability of producing the correct target sequence given the input for all data samples in the training corpus.

\subsection{Data}

We use user-composed e-mails as the training data for Smart Compose models. We tokenize the e-mails into words or wordpieces. Further, we also include other contextual information like:
\begin{itemize}
\item Previous e-mail in case the composed e-mail was a response.
\item Subject of the e-mail.
\item Date and time of the composed e-mail.
\begin{itemize}
\item These features help the model suggest appropriate responses like \textit{Good morning} and \textit{Good evening} based on the time of the day and \textit{Happy new year} and \textit{Happy Thanksgiving} based on month of the year. They are used as discrete features for our model.
\end{itemize}
\item Locale of the user composing the e-mail.
\begin{itemize}
\item 
This feature helps the model distinguish between different spellings like \textit{behavior} and \textit{behaviour} between en-US and en-GB locales.
\end{itemize}
\end{itemize}
We pre-process the data similar to that described in \cite{Kannan16}, namely
\begin{itemize}
\item Language detection: The language of the message is identified and messages outside of the language for which the model is being built are discarded.
\item Segmentation/Tokenization: Sentences boundaries are identified for the content features. Further, sentences are broken into words and punctuation marks. 
\item Normalization: Infrequent words and entities like personal names, URLs, e-mail addresses, phone numbers etc. are replaced by special tokens.
\item Quotation removal: Quoted original messages and forwarded messages are removed. 
\item Salutation/close removal: Salutations like Hi John and closes such as Best regards, Mary are removed.
\end{itemize}

All the experiments in this section were conducted on English data. After the pre-processing steps, there are about 8 billion English messages in our data set. We split the data set into 80\% for training and 20\% for test. For English, we use word-level models. Our vocabulary contains the most frequent 50k English words.

\subsection{Model Architecture}
\label{sec:model_architecture}



At its core, Smart Compose is a sequence prediction task, so a key aspect of our work is comparing and understanding state-of-the-art sequence generation models in this context.
One important design goal we take great care of is how to properly take advantage of contextual information for more adaptive and accurate suggestion generation. The contextual information we consider includes subject, previous mail body, date, time and locale, which provide sufficient cues for the model to be adapted to the current user composition environment.

For this purpose, we discuss three approaches in this section. In all these approaches, the conditional inputs, including tokenized e-mail content, categorical features such as date, time and locale, are fed to the language model through an embedding layer.

\begin{figure}[h]
\begin{subfigure}{.24\textwidth}
  \hspace*{-.5cm}
  \centering
  \includegraphics[height=4.4cm]{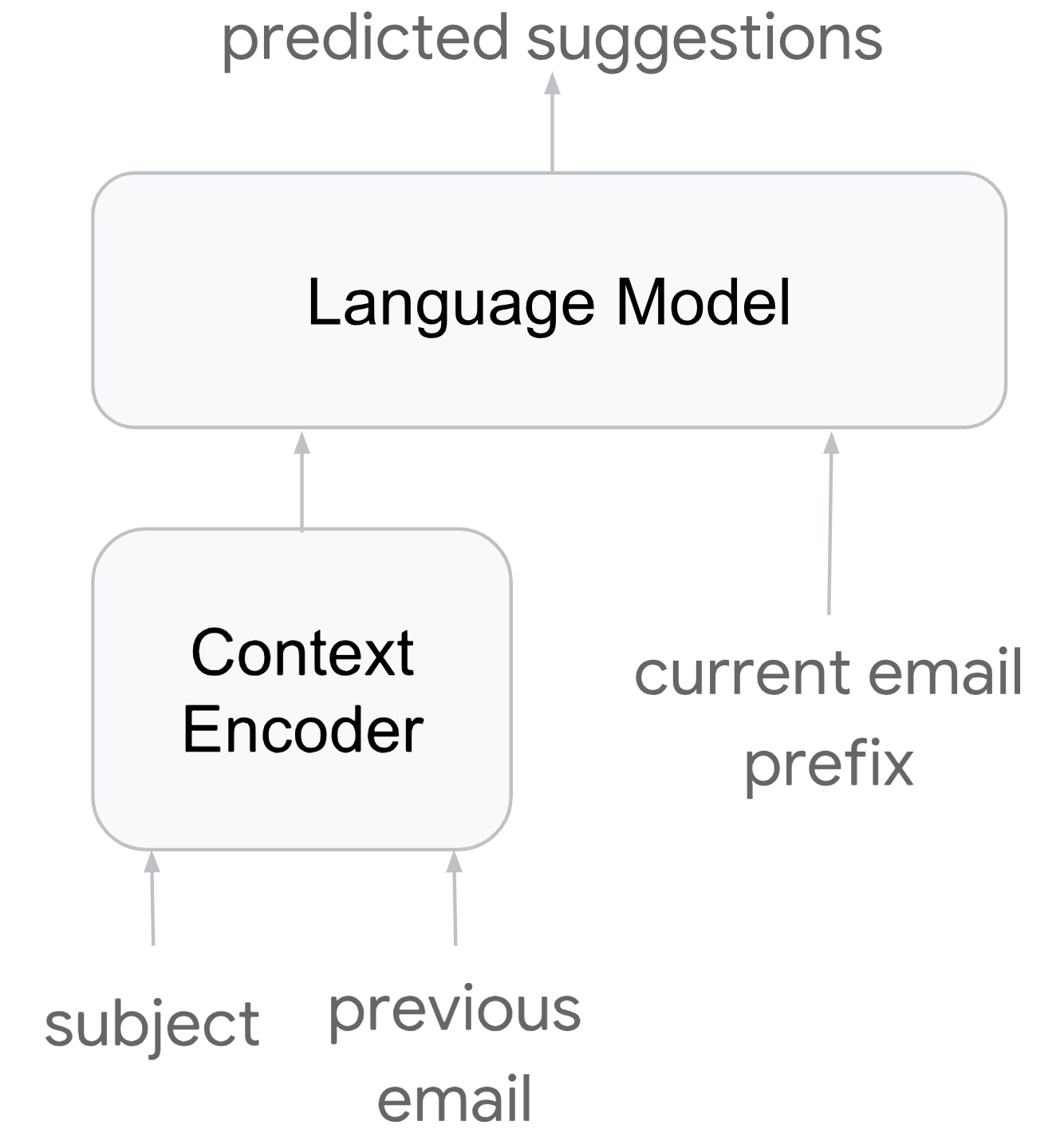}
  \caption{LM-A}
  \label{fig:lma}
\end{subfigure}%
\begin{subfigure}{.26\textwidth}
  \hspace*{-.6cm}
  \centering
  \includegraphics[height=4.4cm]{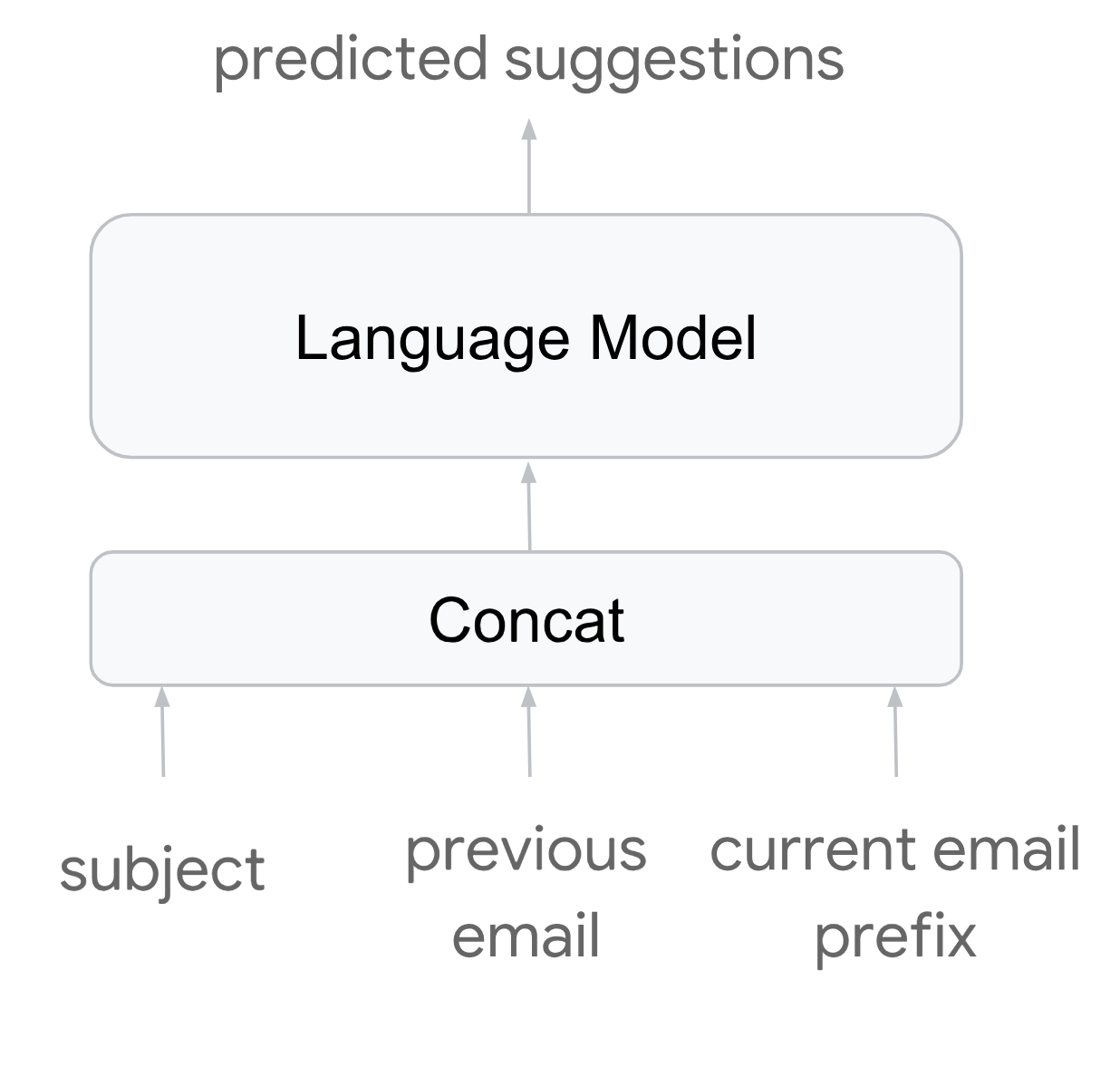}
  \caption{LM-B}
  \label{fig:lmb}
\end{subfigure}
\caption{Smart Compose language generation as language model tasks.
}
\label{fig:lm}
\end{figure}

\begin{itemize}

\item \textit{Language Model A (LM-A).} In this approach (Figure~\ref{fig:lma}), we formulate the problem as a language modeling task. The input sequence to the language model is the current e-mail body prefix. We encode the context via a dedicated context encoder and then combine the encodings with the input sequence at each time step. For simplicity, we use the averaged token embeddings as our context encoding for each field, 
which is then concatenated with the e-mail prefix.

\item \textit{Language Model B (LM-B).} In the second approach (Figure~\ref{fig:lmb}), 
we pack the subject sequence, the previous e-mail body sequence and the current body sequence into a single sequence, with a special token to separate different types of fields. This merged sequence is used as the input to the language model. Compared to LM-A, this approach is simpler in model structure, but it also leads to much longer sequence length.

\begin{figure}[h]
\centering
\includegraphics[width=0.35\textwidth]{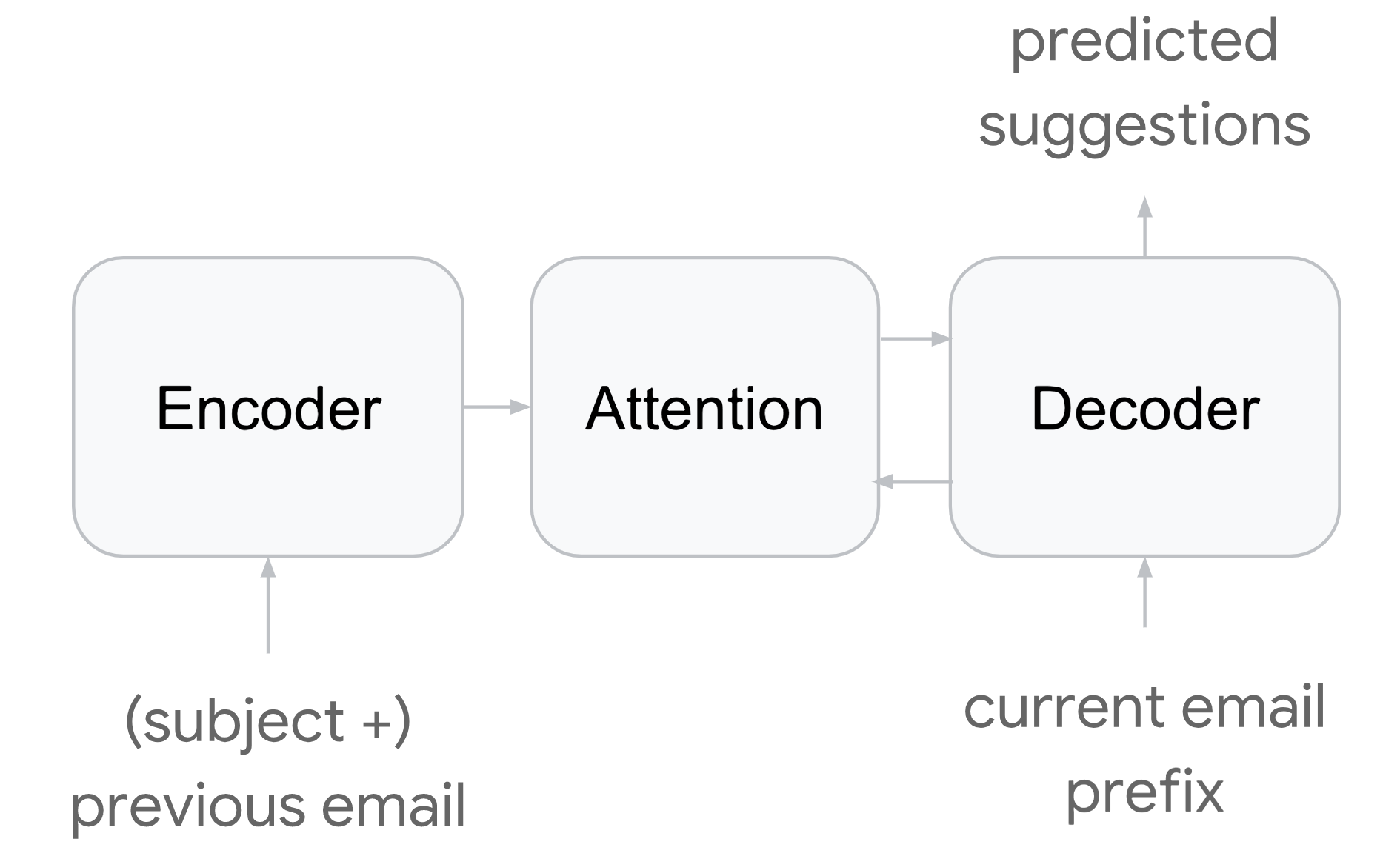}
\caption{Smart Compose language generation as a Seq2Seq task. 
}
\label{fig:s2s}
\end{figure}

\item \textit{Sequence-to-Sequence Model (Seq2Seq).} In this approach (Figure~\ref{fig:s2s}), the problem is formulated as a sequence-to-sequence prediction task similar to neural machine translation, where the source sequence is the concatenation of the subject and the previous e-mail body (if there is one), and the target sequence is the current e-mail body. Through the attention mechanism, a Seq2Seq model has the potential advantage of better understanding the context information.

\end{itemize}

\subsection{Triggering}

At inference time, we feed in necessary context fields (subject, previous e-mail body, etc.) to the model and use a beam search procedure to generate n best suggestions.

We do this by maintaining a heap of m best candidate sequences. At each beam search step, new candidates are generated by extending each candidate by one token and adding them to the heap. Specifically, for each candidate sequence, we use the output of the softmax to get a probability distribution over the vocabulary, select the top k most likely tokens and add the k possible extensions into the heap. We do this for all candidate sequences. At the end of the step, the heap is always pruned to only keep m best candidates. Each candidate sequence is considered complete when a sentence punctuation token or a special end-of-sequence (<EOS>) token is generated, or when the candidate reaches a predefined maximum output sequence length. Upon completion, a candidate sequence will be added to the set of generated suggestions. The beam search ends when no new candidate sequences are added.

In practice, we provide only the top suggestion to users only when the model is "confident" enough about that suggestion. We use a length-normalized log conditional probability as the \textit{confidence score} of each suggestion sequence and define a triggering threshold based on a target triggering frequency/coverage. This confidence score is also used as the value for the ordering within the candidate heap during beam search.

\subsection{Evaluation}

We evaluate the models using the following two metrics:
\begin{itemize}
\item \textbf{Log Perplexity} is a typical measure used for language model evaluation~\cite{Jelinek97}. It is calculated as
\begin{equation}
\textrm{Log Perplexity}(x) = -\sum_x{p(x)\log p(x)}
\end{equation}
where $x$ is the ground truth label and $p(x)$ is the model. The perplexity metric measures how well a model fits the data. A model with lower perplexity assigns higher probabilities to the true target tokens, thus is expected to be more superior for prediction.

\item \textbf{ExactMatch@N}\cite{D16-1264} measures, for a predicted phrase that is N words long, the percentage of predicted phrase that exactly matches the first N words in the ground truth text. This metric allows us to measure how each model performs at different suggestion length. In this paper, we report an averaged ExactMatch number by weighted averaging the ExactMatch for all the lengths up to 15. To ensure a fair comparison, we first select model-specific triggering confidence thresholds to make sure all models have the same coverage, and then compute the ExactMatch results out of triggered suggestions.

\end{itemize}

\subsection{Experiments}

In each of the approaches described in Sec \ref{sec:model_architecture}, we studied both Recurrent Neural Networks (RNN) and Transformer \cite{Vaswani2017}-based model architectures and explored multiple model size variations.

Our RNN-based models use LSTM \cite{Hochreiter1997} layers, with residual connections \cite{He2015} between consecutive layers. We use the Adam optimizer \cite{Kingma2014} and adjust the learning rate according to the schedule proposed for RNMT+ models in \citet{Chen2018}. We also apply uniform label smoothing \cite{Szegedy2015}.
For Transformer models, we follow the optimizer settings and learning rate schedule described in \citet{Vaswani2017}. During training, we also use adaptive gradient clipping: a training step is discarded completely if the log of the gradient norm value exceeds four standard deviations of its moving average. We use synchronous training \cite{Chen2016} for all models. Each of the experiments presented in this section were trained on 32 NVIDIA P100 GPUs. 

Table~\ref{table:lma_pplx} shows the results of LSTM and Transformer models using LM-A with different model sizes. For comparison, we also provide the result of a baseline LSTM language model (first row in the table) without any context embeddings from subject or previous e-mail. For all LSTM models, the token embedding size is 256. In LM-A, by concatenating step embedding with the average token embeddings from the subject and previous e-mail body, the input dimension to the LSTM is 768. For the Transformer experiments, we use 6 layers and 8 attention heads. We set the embedding size to $\frac{1}{3} d_{model}$ so that the actual input dimension to the Transformer model is $d_{model}$.

\begin{table}[h]
\centering
\caption{Results using the LM-A approach. LSTM-$l$-$h$ refers to an LSTM model with $l$ layers and $h$ hidden units per layer. Transformer-$d_{model}$-$d_{ff}$ is a Transformer model with model dimension $d_{model}$ and feed-forward inner layer dimension $d_{ff}$.} 
\label{table:lma_pplx}
\begin{tabular}{ cc>{\centering\arraybackslash}m{1.1cm}>{\centering\arraybackslash}m{1.4cm}}
 \toprule
Model & $\#$ Params & Training Time (h) & Test Log Perplexity\\
\midrule
LSTM-2-1024 (no context) & 77.7M & 72 & 3.39\\
\midrule
LSTM-2-1024 & 79.8M & 74 & 3.26\\
LSTM-4-1024 & 96.6M & 80 & 3.19\\
LSTM-6-1024 & 113.3M & 120 & 3.18\\
LSTM-2-2048 & 171.9M & 138 & 3.13\\
Transformer-768-2048 & 84.3M & 202 & 3.08\\
Transformer-768-4096 & 103.2M & 212 & 3.03\\
Transformer-768-8192 & 141.0M & 366 & 2.96 \\
Transformer-1536-8192 & 310.2M & 387& 2.90\\
 \bottomrule
\end{tabular}
\end{table}

We observed that for the LSTM-2-1024, simply by joining the averaged embeddings from subject and previous e-mail to the model input reduces the test log perplexity by 0.13, showing that contextual information indeed helps improving model quality. For both LSTM and Transformer models, increasing the model capacity by increasing model width and/or depth yields lower perplexities. At a similar level of model capacity in terms of the number of model parameters, Transformer tends to have better quality than LSTM model. For example, the log perplexity of a 84M-parameter Transformer is lower than that of a 80M-parameter LSTM model by 0.18. 

\begin{table}[h]
\caption{ExactMatch results using the LM-A approach.} 
\label{table:lma_em}
\centering
\begin{tabular}{ >{\centering\arraybackslash}m{2.2cm}>{\centering\arraybackslash}m{0.8cm}>{\centering\arraybackslash}m{0.8cm}>{\centering\arraybackslash}m{0.8cm}>{\centering\arraybackslash}m{0.8cm}>{\centering\arraybackslash}m{0.8cm}}
 \toprule
 \multirow{2}{*}{Model} & \multicolumn{5}{c}{ExactMatch}\\
 \cmidrule{2-6}
  & @1 & @2 & @3 & @5 & Overall\\
 \midrule
LSTM-2-1024 & 84.63\% & 51.88\% & 40.25\% & 31.82\% & 66.99\% \\
LSTM-2-2048 & 84.89\% & 54.35\% & 43.79\% & 36.18\% & 68.31\% \\
Transformer-768-2048 & 83.93\% & 51.97\% & 43.37\% & 41.10\% & 66.94\% \\
Transformer-1536-8192 & 85.33\% & 53.14\% & 47.87\% & 43.92\% & 67.73\% \\
 \bottomrule
\end{tabular}
\end{table}

We take four of the models from Table~\ref{table:lma_pplx} and report the ExactMatch results in Table~\ref{table:lma_em}. We observed that perplexity difference in general translates to the ExactMatch metric, especially on longer suggestions. However, since the majority of suggestions tend to be short, the quality gap in terms of the overall average ExactMatch is less pronounced.

\begin{table}[h]
\caption{Results using the LM-B approach.} 
\label{table:lmb}
\centering
\begin{tabular}{ cc>{\centering\arraybackslash}m{1.2cm}>{\centering\arraybackslash}m{1.5cm}}
 \toprule
Model & $\#$ Params & Training Time (h) & Test Log Perplexity\\
 \midrule
LSTM-2-1024 & 77.7M & 72 & 3.26\\
Transformer-512-2048 & 80.2M & 156 & 3.08\\
Transformer-2048-8192 & 507.0M & 480 & 2.86\\
 \bottomrule
\end{tabular}
\end{table}

Table~\ref{table:lmb} shows our results with both model types using the LM-B approach.
We observed that given the same model type and capacity, LM-A and LM-B lead to very similar model quality as well as convergence time.

\begin{table}[h]
\caption{Results using the Seq2Seq approach.} 
\label{table:s2s}
\centering
\begin{tabular}{ cc>{\centering\arraybackslash}m{1.2cm}>{\centering\arraybackslash}m{1.5cm}}
 \toprule
Model & $\#$ Params & Training Time (h) & Test Log Perplexity\\
 \midrule
LSTM-2-1024 & 179.7M & 77 & 3.09\\
Transformer-1024-8192 & 430.7M & 168 & 2.96\\
BiLSTM-6-LSTM-8-1024 & 452.6M & 198 & 2.98\\
 \bottomrule
\end{tabular}
\end{table}

Table~\ref{table:s2s} summarizes the results with the seq2seq approach. In our baseline seq2seq model, both encoder and decoder have two 1024-dimensional LSTM layers.
We can see that the LSTM seq2seq model outperforms the LSTM language models with similar number of model params, indicating that an encoder-decoder architecture with attention is more effective in modeling the context information.

Along this line of experiments, we also evaluated how well the state-of-the-art seq2seq models from neural machine translation can perform on the Smart Compose task. Specifically, we selected Transformer Big model from \citet{Vaswani2017} and an RNN-based RNMT+ model from \citet{Chen2018}. In Table~\ref{table:s2s}, Transformer-1024-8192 is the Transformer Big model, where both the encoder and the decoder have 6 Transformer layers, with model dimension 1024 and feed-forward inner layer dimension 8192. BiLSTM-6-LSTM-8-1024 is the RNMT+ model, where there are 6 bidirectional LSTM layers in the encoder and 8 unidirectional LSTM layers in the decoder. We followed the implementations of the two models released in Tensorflow-Lingvo\cite{Lingvo}. Without any parameter tuning, both Transformer Big model and RNMT+ perform fairly well on the Smart Compose data, with perplexity on par with each other and better than our LSTM seq2seq model baseline.

\section{Production System} \label{production}
Smart Compose is implemented within a streaming RPC server. Application logic that runs in a higher-level Gmail service determines whether a message is eligible for Smart Compose and thus, whether or not to initiate a streaming session with the Smart Compose server. With each new keystroke the user types, new requests are presented to the server.

\subsection{Life of a Smart Compose Request}
Each Smart Compose request is composed of a sequence of prefix encoding steps and beam-search steps. Each of these steps involves computing a single time-step inference of the language model. During the prefix-encoding phase, the goal is to obtain the final hidden states of the language model, which represent the full encoding of the prefix. The beam-search phase then explores hypothetical extensions of the prefix, keeping track of plausible candidate suggestions.

\subsubsection{Context encoding}
The embedding look-ups for features like the previous message body, e-mail subject, date/time etc. are computed once and concatenated to the token embeddings that are fed every time-step in the subsequent phases.
    
\subsubsection{Prefix encoding steps}
Each complete prefix token is fed into the language model along with the previous hidden state output. Since a user has a sticky assignment to a server, much of this computation can typically be skipped by using cached results from previous requests.  Thus, the cost of prefix encoding is typically proportional to how much the user has typed since the last request.
 
\subsubsection{Beam Search steps}
If the prefix ends mid-word, the first step of beam search starts constrained by only feasible completions of the prefix. In this manner, we can do partial word completions, although our model has a word-level vocabulary. Application parameters control the maximum number of steps to run beam search and valid tokens to end consideration of a beam. For instance, if we do not wish to suggest beyond a sentence, the period token would indicate a completed beam. Additionally, tokens that are offensive or otherwise undesirable for product reasons can be dropped at this stage.

\subsection{Balancing Quality and Inference Latency}

Table~\ref{table:lma_latency} summarizes the relative latency results of beam search for the previously mentioned four models in Table~\ref{table:lma_em}. We show the latency per decoding step as well as latency per suggestion, averaged over all samples in our test set. All numbers are compared against the LSTM-2-1024's per-step latency. For per-suggestion latency, we further divide all suggestions into three buckets based on suggestion length (for example, the bucket "1-5" groups the suggestions of length between 1 and 5) and show both bucket-based latency and overall latency.

\begin{table}
\caption{Relative beam search latency using the LM-A approach (measured on CPU). All numbers are compared against the LSTM-2-1024's per-step latency.}
\label{table:lma_latency}
\centering
\begin{tabular}{ >{\centering\arraybackslash}m{2.0cm}ccccc}
 \toprule
 \multirow{2}{*}{Model} & \multirow{2}{*}{Step} & \multicolumn{4}{c}{Suggestion}\\
 \cmidrule{3-6}
  & & 1-5 & 6-10 & 11-15 & Overall \\
 \midrule

LSTM-2-1024           & 1x    & 3.91x  &  8.51x & 12.53x & 9.51x\\
LSTM-2-2048           & 2.29x & 8.78x  & 19.60x & 29.13x & 20.87x\\
Tranformer-768-2048   & 2.22x & 10.78x & 17.58x & 28.27x & 19.71x\\
Transformer-1536-8192 & 5.42x & 25.51x & 40.87x & 69.84x & 37.96x\\
 \bottomrule
\end{tabular}
\end{table}

Within the same model type, increasing the model size naturally leads to latency increase. Note that despite the quality advantage of Transformer model over a similar-sized LSTM model, the average decoding latency of Transformer is much worse than LSTM on both per-step and per-suggestion basis. This is due to keeping track of self-attention keys and values from all previous decoding steps for all the layers of the model. As the number of decoding steps increases, each new step becomes more expensive, which also explains the growing latency gaps between Transformer and LSTM models as the suggestions get longer.


Despite the superior quality from Transformer-based Language Models and more sophisticated seq2seq models, due to the strict production latency constraint and very high request volume, we conclude that LM-A is the most production-appropriate model. We notice that the quality gap between the RNN-based model and Transformer based model is less evident in the ExactMatch metric than in the log perplexity metric, the former of which is more important to production. We hypothesize that this is because the improvement in the log perplexity metric is mostly in places where the model is relatively low in confidence and Smart Compose is unlikely to be triggered.

Table~\ref{table:suggestions} shows some of the suggestions generated by our production model. In many cases, the model is able to capture well the context from subject and previous email, despite the simplistic approach of using averaged embeddings to encode each field.

\begin{table}
\centering
\caption{Example suggestions from production Smart Compose model. In the last column, the current e-mail prefix is separated from the generated suggestion by a `+' sign.} 
\label{table:suggestions}
\begin{tabular}{ >{\centering\arraybackslash}m{1.1cm}>{\centering\arraybackslash}m{2.3cm}>{\centering\arraybackslash}m{4cm}}
 \toprule
Subject & Previous e-mail & Current e-mail + Suggestion \\
\midrule
& Thank you! & Y + \textbf{\textit{ou're welcome}} \\
& Do you want tea? & Y + \textbf{\textit{es please}} \\
Meet & & Look f + \textbf{\textit{orward to seeing you}} \\
Thursday & Thursday is great! & I will work on T + \textbf{\textit{hursday}} \\
Tuesday & Tuesday is great! & I will work on T + \textbf{\textit{uesday}} \\
 \bottomrule
\end{tabular}
\end{table}
    
\subsection{Accelerating Inference}
In order to meet product requirements at scale, it was clear that using hardware accelerators would be necessary. Initially, the first-generation tensor processing unit (TPU) \cite{TPUv1} was used to accelerate matrix-multiplications and activation functions.  However, some unsupported operations such as SoftMax would still be performed on the CPU host.  Subsequently, inference was moved to the second-generation "Cloud TPU" \cite{TPUv2} which has hardware support for a larger amount of Tensorflow operations. The XLA just-in-time compiler \cite{XLA} was used to optimize the graph of operations and fuse operations appropriately for targeting the Cloud TPU hardware.  Thus, most of the operations are executed on the TPU accelerator, whereas the host CPU takes care of application logic, caching and beam search book-keeping.  In Table~\ref{table:inference}, the 90th percentile request latency and total throughput of the various platforms is shown.


\begin{table}
\centering
\caption{Relative request latency and relative throughput comparison to CPU baseline.} 
\label{table:inference}
\begin{tabular}{ ccc }
 \toprule
Platform & Rel. Latency & Rel. Throughput \\
 \midrule
CPU   &   1x  &   1x  \\
TPUv1 & 0.16x & 13.3x  \\
TPUv2 & 0.10x  & 93.3x\\
 \bottomrule
\end{tabular}
\end{table}

To fully take advantage of the hardware accelerator and to increase the maximum throughput,
we batch work across requests. We can batch both prefix encoding steps and beam search steps together for the same language model.

\subsection{Accelerating Training}
Production Smart Compose models are trained using Cloud TPU accelerators in a "quarter-pod", 64-chip setup. Using the same hardware platform for training and serving provides a convenient consistency as there are no longer training/serving skews such as training in floating-point but serving with quantization.

\section{Personalization} \label{personalization}



E-mail writing style can vary a lot from person to person. Applying a single language model uniformly to everyone falls short in capturing the uniqueness of the personal writing style and vocabulary. In order to capture the personal mail writing styles without adding too much burden for serving, we train for each user a light-weight language model adapted to the user's personal mail data.

\begin{table}[h]
\centering
\caption{Suggestions illustrating the contribution of each model on one of our testing accounts.} 
\label{table:p13n_suggestions}
\begin{tabular}{ >{\centering\arraybackslash}m{1.8cm}>{\centering\arraybackslash}m{5cm}}
 \toprule
Model & Current e-mail + Suggestion \\
\midrule
Global-only & Will this work for sm + \textbf{\textit{all group?}} \\
Personal-only & Will this work for sm + \textbf{\textit{artcompose.}} \\
Blended & Will this work for sm + \textbf{\textit{artcompose?}} \\
 \bottomrule
\end{tabular}
\end{table}

\subsection{Model} Due to the large user base of Gmail and strict user data protection policies, personalized model needs to be small, adaptive, easy for serving and periodical re-training. We therefore choose an n-gram language model with Katz-backoff~\cite{katz1987} for this purpose. Compared to a RNN-based model, the Katz n-gram model is easier to train and require much less data to attain a reasonable quality. For efficient storage and inference, we follow common practice and represent the language model in compact weighted finite automata (WFA) format~\cite{Allauzen2003}.
    
The final prediction probabilities are given by the linear interpolation between the personal and global models, computed at each time step of the beam search procedure: \begin{equation} P_{\textrm{final}} = \alpha P_{\textrm{personal}} + (1-\alpha)P_{\textrm{global}} \end{equation} where $\alpha\in[0, 1]$. Ideally, the interpolation weight \(\alpha\) can be context-dependent and estimated by a dedicated model (see for example~\citet{bakhtin2018}). However, in order to avoid introducing too much overhead due to personalization, we keep the model as simple as possible and make use of a constant interpolation weight. The value of the weight is first estimated from offline experiments, then verified by online experiments to ensure suggestion quality improves compared with using global model alone.
    
\subsection{Data} For each user, a dedicated n-gram language model is trained on his or her ``Sent'' e-mails in the past a few months. The personal vocabulary is extracted from the same data set according to word frequency with a minimum number of word occurrence threshold, and the vocabulary size is constrained to be below a maximum threshold. Since there are always some out-of-vocabulary (OOV) words that are not included in either the global or personal vocabularies, we assign OOVs a tiny probability to ensure that the probability of all possible words sum up to one.
    
\subsection{Evaluation} 
\begin{figure}[h]
\centering
\includegraphics[width=0.5\textwidth]{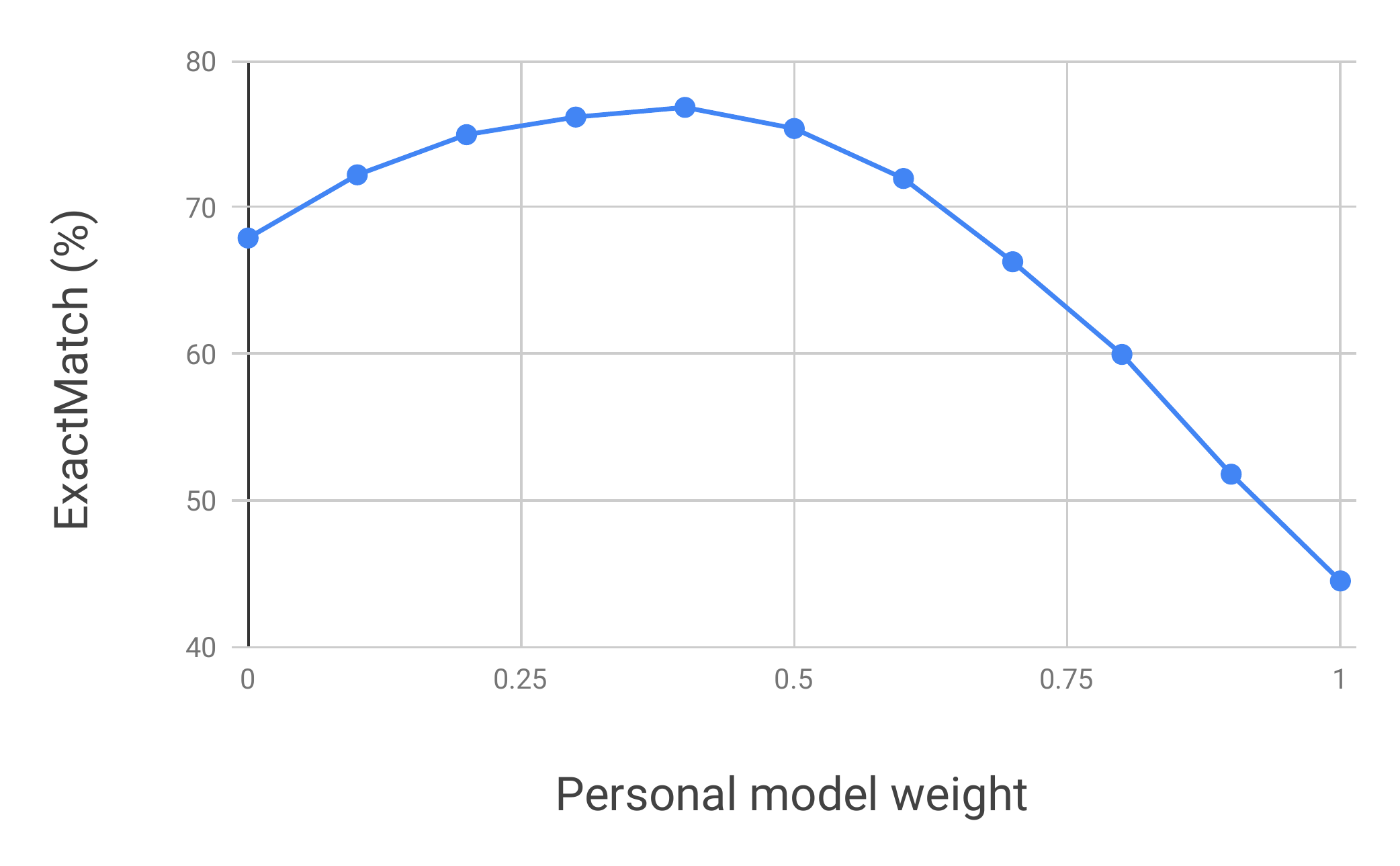}
\caption{Offline evaluation on personal model with different blending weights.}
\label{fig:personal}
\end{figure}

The performance of the personalized model is sensitive to the interpolation weight $\alpha$. Figure~\ref{fig:personal} demonstrates the offline evaluation metric results. We ran the evaluation on real user data for 11 rounds, each round using a different blending weight $\alpha$ (``Global-only'' is the case where $\alpha=0$). For fair comparison, we tried our best to adjust the confidence threshold so that the coverage in different experiments are roughly the same and on par with the global-only model.
    
Overall, the ExactMatch numbers show that personalized models outperform the global model when $\alpha$ is properly set. The performance improves as we increase the $\alpha$ value, peaks at 0.4, and starts to drop as we further tune up the value. This makes sense since the global model alone $(\alpha=0)$ fails to capture personal styles while a pure personal ngram model $(\alpha=1)$ is not as powerful as the global neural language model in making generic predictions. 
After we launched the personalized model to production, we have observed around $6\%$ relative gain in the Click-Through-Rate (CTR) of suggestions and $10\%$ relative gain in ExactMatch. In Table~\ref{table:p13n_suggestions}, we also show some examples comparing the suggestions given by global, personal and interpolated models. It can be seen that while global model only makes general suggestions, with the help of personal model the suggestions become tailor-made for individual users.

\section{Multilingual Model} \label{multilingual}

In extending Smart Compose models to other languages including Spanish (es), French (fr), Italian (it) and Portuguese (pt), we set out to explore the path of multilingual models. Multilingual models have the advantage of being easier to maintain and serve, while allowing low-resource languages to benefit from high-resource languages in terms of model quality. Since word-level multilingual models require a large vocabulary size to account for rare words in each language, it would make the models prohibitively slow to serve within systems with strict latency requirements. We use WordPiece \cite{wordpiece} models and split the tokens into sub-word units (wordpieces). In this way, we can limit the shared wordpiece vocabulary size while still maintaining a low OOV rate for tokens in the individual languages.

There are about 4 billion Spanish messages, 2 billion French messages, 1 billion Italian messages and 2 billion Portuguese messages in our data set. Similar to English, we split the data set of each language into 80\% for training and 20\% for test.

To have a better assessment of the model quality, we compare our multilingual wordpiece model with both monolingual word models and monoligual wordpiece models for each of the four languages. For the experiments conducted in this section, we use a vocabulary size of 50k for all word and wordpiece vocabularies and use the same model architecture (LSTM-2-1024 using LM-A approach) and the same training configurations for all models.

Table~\ref{table:multilingual} summarizes the test log perplexity and ExactMatch results. For all four languages, we observed that monolingual word models yield the lowest perplexities and that monolingual wordpiece models have lower perplexities than multilingual wordpiece models. In terms of ExactMatch, for Spanish, French and Italian, monolingual word models have the best performance, while monolingual wordpiece models outperform multilingual wordpiece models. The quality gaps are displayed most evidently on Spanish, given that Spanish is the highest-resource language among the four languages and thus its quality would likely be impaired the most by using a shared multilingual model. For Portuguese, the multilingual wordpiece model performs better than both monolingual models, showing that it benefits from leveraging more data in similar languages.

Based on an overall consideration of both quality and maintenance requirements, we deployed a multilingual wordpiece model in production for Spanish, French, Italian and Portuguese, with language-specific triggering confidence threshold to obtain similar coverage among the four languages as well as with the English model.

\begin{table}
\caption{Perplexity and ExactMatch results comparing monolingual and multilingual models.} 
\label{table:multilingual}
\centering
\begin{tabular}{ccc>{\centering\arraybackslash}m{1.7cm}}
 \toprule
& Model & ExactMatch & Test Log Perplexity\\
 \midrule
\multirow{3}{*}{es} & monolingual word & 75.90$\%$ & 2.85\\
& monolingual wordpiece & 72.59$\%$ & 2.95\\
& multilingual wordpiece & 71.73$\%$ & 2.98\\
 \midrule
\multirow{3}{*}{fr} & monolingual word & 70.75$\%$ &2.75\\
& monolingual wordpiece & 69.54$\%$ &2.90\\
& multilingual wordpiece & 69.64$\%$ &3.05\\
 \midrule
\multirow{3}{*}{it} & monolingual word & 70.50$\%$ &1.75\\
& monolingual wordpiece & 67.44$\%$ &1.85\\
& multilingual wordpiece & 67.48$\%$ &2.15\\
 \midrule
\multirow{3}{*}{pt} & monolingual word & 69.04$\%$ &3.15\\
& monolingual wordpiece & 72.20$\%$ &3.27\\
& multilingual wordpiece & 72.65$\%$ &3.48\\
 \bottomrule
\end{tabular}
\end{table}


\section{Fairness and Privacy} \label{fairness}
Models trained over natural language data are susceptible to providing suggestions reflecting human cognitive biases \cite{Caliskan}. For Smart Compose models, we found instances of gender-occupation bias when testing our models. For example, when a user was composing an e-mail with text \textit{I am meeting an \textbf{investor} next week}, the models suggested \textit{Did you want to meet \textbf{him}} whereas on typing \textit{I am meeting a \textbf{nurse} next week}, the models suggested \textit{Did you want to meet \textbf{her}}. To limit the impact, we removed any suggestions with a gender pronoun and continue to explore algorithmic approaches to handle bias.

Since Smart Compose is trained on billions of phrases and sentences, similar to the way spam machine learning models are trained, we have done extensive testing to make sure that only common phrases used by multiple users are memorized by our model, using findings from \citet{Carlini2018}.

\section{Future Work} \label{future}

In this work, we have focused on experimenting with recent training techniques and state-of-the-art architectures for sequence prediction tasks. While we demonstrated that these advanced models and techniques greatly improve the suggestion quality of the Smart Compose language generation system, most of them failed to meet our strict latency constraints. In the future, we plan to continue working on two directions:
\begin{itemize}
    \item improving model quality with advanced model architectures and techniques at minimum latency cost;
    \item taking advantage of the latest hardware accelerators and continuously optimizing inference and serving.
\end{itemize}

Specifically, we plan to look into an adapted version of the Transformer model where self-attention is applied locally over a fixed-sized window of steps during decoding, instead of over all previous decoding steps. In this way, we are hoping to maintain the quality improvement of the Transformer model while mitigating the increase in inference latency. Also, inspired by recent work in language model pre-training \cite{dai2015semi,Devlin2018}, we are interested in exploring how similar pre-training techniques would benefit our global English model as well as the multilingual model.

Another line of work worth exploring is to apply variational methods to language modeling. Variational auto-encoders (VAE)~\cite{Kingma14b} allow efficient learning and inference to be performed in a latent variable model. VAEs are able to extract and control the subtle latent information that is not explicitly modelled by frequentist models. Several extensions of VAE to recurrent neural network models have been proposed (see for example~\citet{bowman2015generating}), and they can be readily applied to RNN-based LMs to allow more control over text generation. With the introduction of latent variables, LMs are expected to be able to capture hidden features like topic, sentiment writing style etc. from the text, potentially making Smart Compose suggestions more appropriate and diverse.

\section{Conclusion} \label{conclusion}
In this paper we described Smart Compose, a novel system that improves Gmail users' writing experience by providing real-time, context-dependent and diverse suggestions as users type. We gave detailed accounts of the challenges we faced, including model design, evaluation, serving, and issues like privacy and fairness. We also introduced extended features including personalization and multilingual support, and showed extensive experimental results comparing performances given by different model and serving architectures.


\begin{acks}
We would like to thank Hallie Benjamin, Ulfar Erlingsson, Anna Turner, Kaushik Roy, Katherine Evans, Matthew Dierker, Tobias Bosch, Paul Lambert, Thomas Jablin, Dehao Chen, Steve Chien, Galen Andrew, Brendan McMahan, Rami Al-Rfou, DK Choe, Brian Strope and Yunhsuan Sung for their help and discussions. We also thank the entire Google Brain team and Gmail team for their contributions and support.

\end{acks}

%
\bibliographystyle{ACM-Reference-Format}
\bibliography{kdd}

%
\appendix

\end{document}